\documentclass[letterpaper]{article} % DO NOT CHANGE THIS
\usepackage{aaai25} % DO NOT CHANGE THIS
\usepackage{times}  % DO NOT CHANGE THIS
\usepackage{helvet}  % DO NOT CHANGE THIS
\usepackage{courier}  % DO NOT CHANGE THIS
\usepackage[hyphens]{url}  % DO NOT CHANGE THIS
\usepackage{graphicx} % DO NOT CHANGE THIS
\usepackage{natbib}  % DO NOT CHANGE THIS AND DO NOT ADD ANY OPTIONS TO IT
\usepackage{caption} % DO NOT CHANGE THIS AND DO NOT ADD ANY OPTIONS TO IT
\usepackage{algorithm}
\usepackage{algorithmic}

\usepackage{listings}
\usepackage{amsmath,amssymb,amsfonts}

\usepackage{graphicx}
\usepackage{textcomp}
\usepackage{amsthm}

\usepackage{caption}
\usepackage{subfigure}
\usepackage{multirow}
\usepackage{booktabs}
\usepackage{enumitem}
\DeclareCaptionStyle{ruled}{labelfont=normalfont,labelsep=colon,strut=off} % DO NOT CHANGE THIS
\floatstyle{ruled}
\newfloat{listing}{tb}{lst}{}
\floatname{listing}{Listing}
\title{A Lightweight Sparse Interaction Network for Time Series Forecasting}
\author {
    Xu Zhang\textsuperscript{\rm 1},
    Qitong Wang\textsuperscript{\rm 2},
    Peng Wang\textsuperscript{\rm 1}\thanks{Corresponding author},
    Wei Wang\textsuperscript{\rm 1}
}
\affiliations {
    \textsuperscript{\rm 1}Shanghai Key Laboratory of Data Science, School of Computer Science, Fudan University, Shanghai, China\\
    \textsuperscript{\rm 2}Universite Paris Cite, Paris, France\\
    xuzhang22@m.fudan.edu.cn, \{pengwang5,weiwang1\}@fudan.edu.cn, qitong.wang@u-paris.fr
}
\usepackage{bibentry}
\begin{document}

\maketitle

%  On the one hand, We empirically find the forecasting errors remain at a similar low level until the sparsity of the attention matrix exceeds 90% in popular datasets, e.g., the red numbers shown in the Figure, which is a case study and the sparsity equals (100-$K$)\%. 
% Under such sparsity, we observe that only a small part of the learned attention scores are important for TSF accuracy (most highlighted area in the heatmap of the Figure).
% Hence, the essence of SAM can be regarded as using attention scores to find the important interactions between time steps, thereby capturing temporal dependence. 
% However, considering the most learned lower scores are redundant and ineffective, using SAM to find the small part of important interactions is too computationally expensive. Hence, inspired by Word2Vec, we treat the indices of the attention matrix as a vocabulary and introduce a sparse Bernoulli distribution. Index embeddings are then used to directly predict the latent sparse pairwise interactions between time steps, replacing the standard Q–K–V computation.

\begin{abstract}
Recent work shows that linear models can outperform several transformer models in long-term time-series forecasting (TSF). 
However, instead of explicitly performing temporal interaction through self-attention, linear models implicitly perform it based on stacked MLP structures, which may be insufficient in capturing the complex temporal dependencies and their performance still has potential for improvement.
To this end, we propose a Lightweight Sparse Interaction Network
(LSINet) for TSF task.
Inspired by the sparsity of self-attention, we propose a Multihead Sparse Interaction
Mechanism (MSIM). 
Different from self-attention, MSIM learns the important connections between time steps through sparsity-induced
Bernoulli distribution to capture temporal dependencies for TSF. 
The sparsity is ensured by the proposed self-adaptive regularization loss.
Moreover, we observe the shareability of temporal interactions and propose to perform Shared Interaction Learning (SIL) for MSIM to further enhance efficiency and improve convergence.
LSINet is a linear model comprising only MLP structures with low overhead and equipped with explicit temporal interaction mechanisms. 
Extensive experiments on public datasets show that LSINet achieves both higher accuracy and better efficiency than advanced linear models and transformer models in TSF tasks. The code is available at the link \url{https://github.com/Meteor-Stars/LSINet}.

\end{abstract}


\section{Introduction}
Time Series Forecasting (TSF) is crucial in many scenarios, spanning a wide range of fields such as weather~\citep{angryk2020multivariate_intr_weather}, traffic flow ~\cite{chen2001freeway_traffic}, financial investment~\cite{zhang2024self}, and medical diagnostics~\cite{churpek2016value_intr_medical}. 
The goal is to predict future values by analyzing historical time windows. 
Transformer~\cite{vaswani2017attention} has been popular in TSF tasks due to its powerful capability to capture temporal dependencies~\cite{zhou2021informer,wu2021autoformer,zhou2022fedformer}, but they suffer from quadratic memory and runtime overhead due to self-attention mechanism, which limits their application and scalability. 
Although PatchTST~\cite{nie2022time_patchformer} reduces time complexity and achieves SOTA accuracy by forecasting at the patch level, it still uses the multi-head self-attention under the hood, which is also inevitably computationally expensive~\cite{vijay2023tsmixer}.

 \begin{table}[bt]
 %%\vspace{-0.2cm}
    % \renewcommand{\arraystretch}{0.5}
    \setlength{\tabcolsep}{4pt}
    % {|>{\setlength{\tabcolsep}{3pt}}c|c|c|}
    \centering

    %\vspace{-0.2cm}
    % \begin{tabular}{c|c|p{20pt}p{20pt}|cc|cc|cc|cc|cc}
    % \small
    { \small
    \begin{tabular}{c|ccc|ccc} 
        \hline
        \multirow{2}{*}{\shortstack{Datasets/\\ Methods}} &  \multicolumn{3}{c|}{Weather ($bs$=64)} &  \multicolumn{3}{c}{Electricity ($bs$=16)}\\
         &  \textit{Epo.$^T$} &\textit{Infer$^T$}& \textit{Mem.}  &  \textit{Epo.$^T$} &\textit{Infer$^T$}  & 
         \textit{Mem.}    \\ 
        \midrule[0.5pt]
        \multicolumn{1}{c|}{PatchTST} &59.98 &32.73 &5034 &377.25  &200.88 &16128\\
        \midrule[0.5pt]
        \multicolumn{1}{c|}{CI-TSMixer} &52.91 &32.67 &4276 &323.06  &165.56 &13132\\  
        \midrule[0.5pt]
        \multicolumn{1}{c|}{LSINet} &18.92 &17.77 &2940 &91.25  &82.53 &6936\\  
        \midrule[0.5pt]

    \end{tabular}}%small
        \caption{Efficiency comparison of transformer model PatchTST and linear models (CI-TSMixer and our LSINet) on training time \textit{Epo.$^T$}(s/epoch), inference time \textit{Infer$^T$} and memory \textit{Mem.}(MB). The input length for linear models is 1024 while for PatchTST is 512. $bs$ represents batch size.}
            \label{tab:eff_comp_linear_transf}
%\vspace{-0.5cm}
\end{table}

Recently, linear models (consisting of only MLP structure) that do not involve self-attention mechanisms have received emerging attention in TSF tasks due to their competitive accuracy and significant efficiency advantage~\cite{zeng2023transformers_linear,vijay2023tsmixer}. 
As shown in Table~\ref{tab:eff_comp_linear_transf}, despite 
linear models LSINet (ours) and CI- TSMixer~\cite{vijay2023tsmixer} address a longer input length of 1024, their efficiency is higher than transformer model PathchTST that addresses a shorter input length of 512.
One characteristic of highly efficient linear models is that they implicitly perform temporal interaction for TSF~\cite{vijay2023tsmixer}. 
However, the Self-Attention Mechanism (SAM) that generates explicit temporal interaction has been proven powerful in modeling temporal dependencies in the TSF field~\cite{wu2021autoformer,nie2022time_patchformer,li2019enhancing,zhou2022fedformer} and implicit temporal interaction of linear model may be insufficient in capturing the complex temporal dependencies for TSF. 
% Hence, the performance of linear model still has potential for improvements.
Hence, we try to design a linear model equipped with efficient and explicit temporal interaction mechanisms and further explore the potential of the linear model in the TSF task.

\begin{figure}[bt]
%\vspace{-0.2cm}
\centerline{\includegraphics[width=1\linewidth]{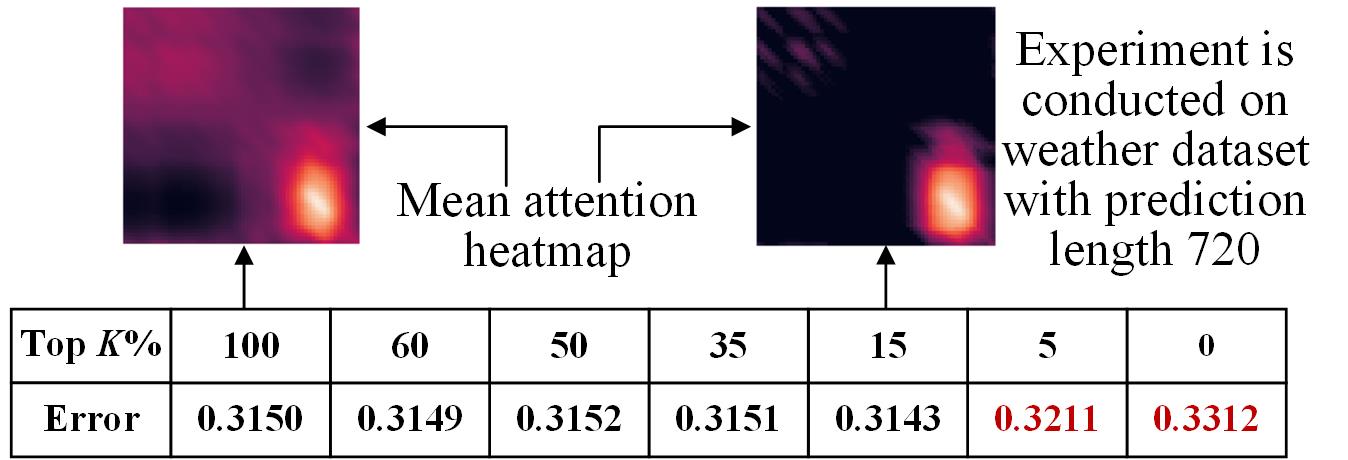} }
%\vspace{-0.2cm} %
\caption{Forecasting error changes after retaining top $K$\% higher self-attention scores (lower is set to zero) on the well-trained PatchTST model and test set of weather dataset. }
\label{fig:moti_weather0619}
%\vspace{-0.6cm}
\end{figure}

Self-Attention Mechanism (SAM) is proved to have the characteristic of \textit{sparsity}~\cite{zhou2021informer} in the TSF task.
We also empirically find the forecasting errors remain at a similar low level until the sparsity of the attention matrix exceeds 90\% in popular datasets, e.g., the red numbers shown in Figure~\ref{fig:moti_weather0619}, which is a case study and the sparsity equals (100-$K$)\%. 
Under such sparsity, we observe that only a small part of the learned attention scores are important for TSF accuracy (most highlighted area in the heatmap of Figure~\ref{fig:moti_weather0619}).
Hence, the essence of SAM can be regarded as using attention scores to find the important interactions between time steps, thereby capturing temporal dependence. 
However, considering the most learned lower scores are redundant and ineffective, using SAM to find the small part of important interactions is too computationally expensive.

Inspired by the understanding of sparsity towards SAM, we propose a Sparse Interaction Mechanism (SIM) to find the 
sparse and important interactions, reducing computational and memory costs.
\textbf{Firstly}, we propose to make the model learn the 0-1 connection matrix $\mathbb{C} \in N\times N$ based on the parameterized Bernoulli distribution. $\mathbb{C}$ contains important connections between time steps that may be effective for capturing temporal dependence. 
$N$ is the total number of time steps and $c_{ij}=1$ in $\mathbb{C}$ denotes an important interaction between $i$-th and $j$-th time steps. 
The normal Bernoulli distribution can roughly learn important connections between time steps.
\textbf{Secondly}, we further propose a self-adaptive sparse regularization loss to add the sparsity into Bernoulli distribution (called sparsity-induced one), thereby limiting the sparsity of $\mathbb{C}$, which makes the model further refinedly learn the important connections between time steps (most highlighted interaction area in the heatmap of Figure~\ref{fig:moti_weather0619}) and drop the redundant ones for more accurately capturing temporal dependence. 
\textbf{Thirdly,} we design Multihead SIM (called MSIM) for increasing interaction diversity.

Besides, unlike the TSF transformer models, which learn self-attention matrices (interaction patterns) for each sample and each variable.
We learn shared connection matrices for all samples and variables (called Shared Interactions Learning, SIL) based on the MSIM to further enhance efficiency and make the model easy to converge.
The key motivation is that temporal interaction patterns tend to occur repeatedly in a time series scenario. 
For example, in weather forecasting tasks, the daily temperature always rises gradually from night to noon and then begins to fall from the afternoon to the evening, leading to repeated temporal patterns.

\begin{figure}[bt]
%\vspace{-0.2cm}
\centerline{\includegraphics[width=\linewidth]{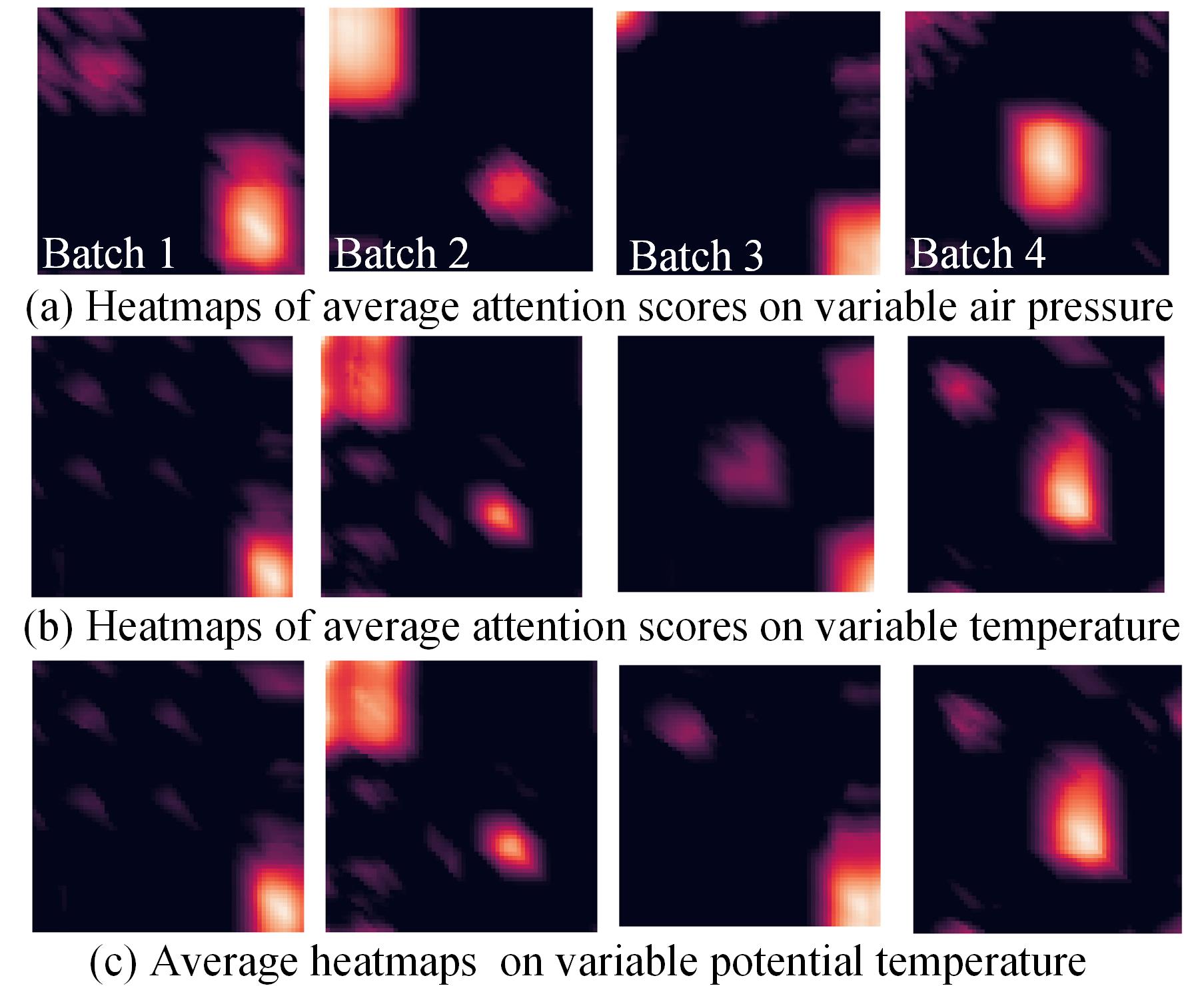} }
%\vspace{-0.4cm} %调整的是Figure字样和图片的距离
\caption{Illustration of repeated interaction patterns in self-attentions based on the test set of weather dataset and PatchTST model. Each batch consists of 64 test samples.}
\label{fig:moti_reperat_pattern0619v2}
%\vspace{-0.5cm}
\end{figure}

We also empirically observed the repeated interaction patterns in the self-attention heatmap, as shown in Figure~\ref{fig:moti_reperat_pattern0619v2}.
On the weather dataset, through the heatmap of a single variable (e.g., air pressure in Figure~\ref{fig:moti_reperat_pattern0619v2}(a)) with different sample batches, we observe important temporal interactions (the most highlighted areas) always repeatedly occur along the diagonal of the heatmap. 
This is reasonable because the test samples are constructed through a sliding window. The historical window's start and end time steps continuously change as the sliding window moves, leading to similar interactions reoccurring along the diagonal of the heatmap. 
Nevertheless, this may lead to the failure of Shared Interactions Learning (SIL) because the historical window is dynamic and the shared connection matrix is static.
To address this, we propose the time-invariant component to mix temporal features into different temporal positions. 
Hence, for any historical time window, the model can capture time dependence through the learned shared connections.

Moreover, through the heatmaps of different weather variables (Figure~\ref{fig:moti_reperat_pattern0619v2}(a), (b), and (c)), we observe that the temporal interactions between variables are also repetitive. 
This is reasonable because the variables are highly correlated (e.g., variable of air pressure and temperature) and the temporal pattern of one variable is also effective for another one, i.e., temporal pattern also occurs repeatedly among different variables. 
The above observations and analyses make us realize that learning interaction patterns for every sample and variable may not be essential because their patterns are repetitive and redundant, which may increase the model convergence's difficulty and lead to expensive and unnecessary model costs. 
Hence, we propose to perform Shared Interactions Learning (SIL) for samples and variables.

In summary, this paper proposes a Lightweight Sparse Interaction Network (LSINet) for time series forecasting. 
To our knowledge, we are the first to explore a linear model equipped with efficient and explicit temporal interaction mechanisms. Our contributions are as follows:
\begin{itemize}
    \item We propose a Lightweight Sparse Interaction Network (LSINet) for Time Series Forecasting (TSF). LSINet comprised two key designs of Multihead Sparse Interaction Mechanism (MSIM) and Shared Interactions Learning (SIL). It bypasses the computation of self-attention scores and learns shared sparse interactions to capture temporal dependence with low overhead.

    \item In MSIM, we propose to learn the important connections between time steps through sparsity-induced Bernoulli distribution (realized by the proposed self-adaptive regularization loss), for capturing temporal dependencies.

    \item We conduct extensive experiments on real-world datasets and achieve advanced forecasting accuracy and efficiency. Moreover, efficiency experiments have proven that our method has faster speed and lower memory overhead compared to existing transformer and linear models.
\end{itemize}

%\vspace{-0.1cm}
\section{Related Work}
\label{sec:related_work}
\label{sec:related_work}
\textbf{Time Series Forecasting (TSF)}

% The TSF algorithms can be categorized into the following types: 
\textbf{(i) Transformer models for TSF.} 
Most transformers learn self-attention at the finest time point level~\cite{zhou2021informer,wu2021autoformer,zhou2022fedformer} while PatchTST~\cite{nie2022time_patchformer} learns self-attention at the patch level, achieving advanced performance in long-term TSF tasks. 
Recently, Scaleformer~\cite{shabani2022scaleformer} proposes a multi-scale structure by iteratively refining forecasted time series at increasingly fine-grained scales. Pathformer~\cite{chen2024pathformer} proposes a path search module to adaptively extract and aggregate multi-scale temporal features for TSF.

\textbf{(ii) Linear models for TSF.} 
% DLinear~\cite{zeng2023transformers_linear} challenges transformer models and outperforms most ones except PatchTST in long-term TSF tasks. 
% For improving DLinear, 
% TSMixer~\cite{chen2023tsmixer} performs TSF by stacking multi-layer perceptrons (MLPs). 
DLinear~\cite{zeng2023transformers_linear} and TSMixer~\cite{chen2023tsmixer} are popular baselines for TSF, which use simple linear layers.
Then, CI-TSMixer~\cite{vijay2023tsmixer} further refines the structure of TSMixer and designs the reconciliation heads with gated attention for TSF, which is proven to outperform PatchTST. 
Recently, TimeMixer~\cite{wangtimemixer} applies the trend-seasonal decomposition mixing at multi-scale level but the operations show high computational cost.
In summary, current linear models haven't explored explicit temporal interaction designs like self-attention for better temporal modeling, which is worth further research.

Finally, Large Language Models (LLMs) are also used in TSF, especially for few-shot learning~\cite{jin2023time,zhou2023one,liu2024taming,alnegheimish2024large,dasdecoder}. 
Nonetheless, researchers also express worry about using LLMs for TSF~\cite{tan2024language}.

\textbf{Multihead Sparse Interaction Mechanism (MSIM) and Sparse Self-attention}

Sparse self-attention such as  \textit{ProbSparse} of Informer~\cite{zhou2021informer}, focuses on enhancing the computational efficiency of the self-attention mechanism by allowing each key to only attend to sparse queries. 
MSIM is different from \textit{ProbSparse} and it doesn't involve computation of queries and keys. 
Because MSIM does not use attention scores to find important interactions between time steps.
In contrast, MSIM finds it through the sparsity-induced Bernoulli distribution. 

% \subsection{Preliminary} 
% \textbf{Problem Formulation}
% Given a historical multivariate time series instance $\mathcal{X}_h=[x_1,x_2,...,x_n] \in \mathbb {R}^{n \times c}$ with the length of $n$, TSF tasks aim to forecast the future $m$ steps $\mathcal{X}_f=[x_{n+1},x_{n+2},...,x_{n+m}] \in \mathbb {R}^{m \times c}$ for all $c$ variables. 

% \textbf{Time Series Patching.} Unlike text in natural language, the semantics of a single time point in TS data is sparse. 
% Dividing the time series into multiple patches (sub-sequences) has been demonstrated more effective for self-attention interactions than using single time points as basic time steps~\cite{nie2022time_patchformer}.
% In this work, we also use time patch for sparse temporal interactions.

% \begin{figure}[bt]
% %\vspace{-0.4cm}
% % \setlength{\abovecaptionskip}{0.1cm} 
% \centerline{\includegraphics[width=0.9\linewidth]{Figure/patching_illustrate0713.jpg} }
% %\vspace{-0.4cm}
% \caption{TS patching with patch length $L$ equals stride $K$.}
% %\vspace{-0.6cm}
% \label{fig:patching_illustrate2}
% \end{figure}

%\vspace{-0.1cm}
\section{A Lightweight Sparse Interaction Network (LSINet) for Time Series Forecasting (TSF)}
\begin{figure*}[bt]
%\vspace{-0.2cm}
% \setlength{\abovecaptionskip}{0.1cm} 
\centerline{\includegraphics[width=1\linewidth]{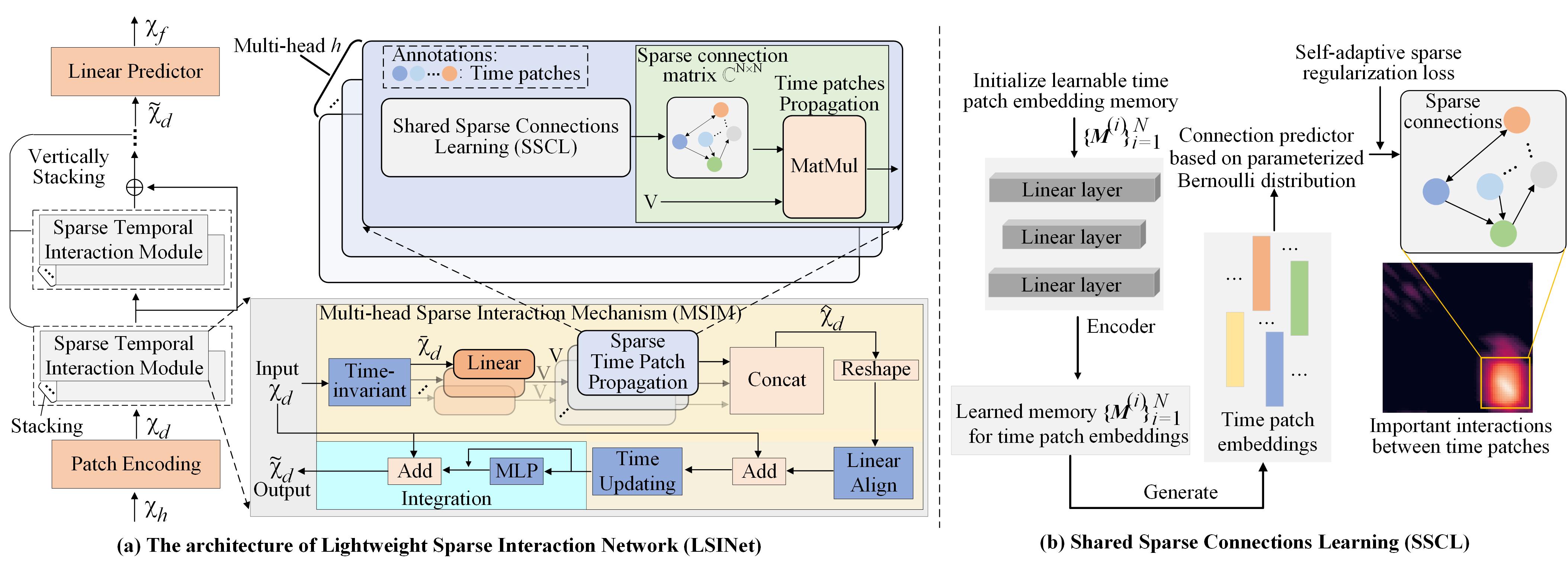} }
%\vspace{-0.4cm}
\caption{LSINet and corresponding components.}
%\vspace{-0.7cm}
\label{fig:framework_LSTINet}
\end{figure*}

The architecture of LSINet is shown on the left of Figure~\ref{fig:framework_LSTINet}. 
The historical TS $\mathcal{X}_h$ is first sent into the Patch Encoding module, which includes operations of Instance Normalization, TS patching, TS embedding, and position embedding. 
After $\mathcal{X}_h$ undergoing these operations, the Patch Encoding module produces $\mathcal{X}_d$, which is further sent into the Sparse Temporal Interaction (STI) module. 
In STI, the time-invariant component first mixes the temporal patch feature of $\mathcal{X}_d$ into different temporal positions. 
Thus, the subsequent shared connection matrix in the Multihead Sparse Interaction Mechanism (MSIM) can be used for capturing temporal dependence through Sparse Time Patch Propagation and Time Updating components in Figure~\ref{fig:framework_LSTINet}(a). 
Specifically, time patch propagation is finished based on multiplication between the temporal input $V$ and sparse connection matrix $\mathcal{C}$, which is obtained by the Shared Sparse Connections Learning (SSCL) component (Figure~\ref{fig:framework_LSTINet}(b)). 
Then, Time Updating is finished by using an MLP to project the temporal patch dimension. 
Next, another MLP is used to integrate the interacted temporal feature while residual connections are used to ensure the STI modules avoid performance degradation when stacked. 
Finally, the output of STI is flattened and sent to a linear predictor for forecasting. The whole LSINet is formed by the stacked STI modules.

\subsection{Patch Encoding}
We first apply Instance Norm (a normalization technique)~\cite{DBLP:conf/iclr/KimKTPCC22} for historical input $\mathcal{X}_h$ to address the distribution shift between training and testing data. 
Then, we transform $\mathcal{X}_h$ with length $n$ into multiple patches with patch length $L$ and stride $K$. 
$L$ is the sequence length of each patch
while $K$ controls the non-overlapping region between consecutive patches. 
If $L$ equals $K$, there will be no overlapping parts between consecutive patches.
Following~\cite{nie2022time_patchformer}, we pad $K$ repeated numbers of the last value to the end of the original sequence before patching. 
Hence, based on patch length $L$ and stride $K$, the patch function $Patch (\mathcal{X}_h,L,K)$ divides original TS $\mathcal{X}_h$ into $N$ patches, denoted by $\mathcal{X}_p$. Each patch in $\mathcal{X}_p$ contains $L$ time steps. The number of patch $N$ is computed by:
% %\vspace{-0.1cm}
 \begin{equation}
 %\vspace{-0.1cm}
\label{equ:patch_cons}
N=\lfloor (n-L)/K \rfloor +2
%\vspace{-0.05cm}
\end{equation}
where $\lfloor \cdot \rfloor$ denotes rounding down.

Then, we perform patch projection, which uses a trainable linear layer to map the patches into high-dimensional space with size $D$ for better temporal interactions.
Since our interaction mechanism also doesn't consider the position information, we add learnable position encoding~\cite{nie2022time_patchformer} for each $\mathcal{X}_p$ after patch projection.
The patch projection and position encoding can be formulated as follows:   
 \begin{equation}
\label{equ:linear_proj_pos}
\mathcal{X}_d= W_{p}\mathcal{X}_p+W_{pos}
\end{equation}

where $W_p \in \mathbb {R}^{D \times L}$ and $W_{pos} \in \mathbb {R}^{D \times N}$. 

\subsection{Sparse Temporal Interaction (STI) Module}
The key design of STI is the Multi-head Sparse Interaction
Mechanism (MSIM), which depends on the Shared Sparse Connections Learning (SSCL) component (Figure~\ref{fig:framework_LSTINet}(b)). 

\textbf{Shared Connection Matrix in SSCL}
We observe that the interaction pattern (connection matrix) of TS data tends to occur repeatedly along the time axis. 
Moreover, the variables in multivariate TS are highly correlated, e.g., the variables of air pressure and temperature. 
Hence, the interaction pattern of one variable is also effective for another one, i.e., the interaction pattern also occurs repeatedly among different variables. 
Inspired by the above analysis, learning the interaction pattern individually for each variable and sample may lead to unnecessary computational overhead. 
Thus, we propose to learn the shared interaction pattern (connection matrix) for samples and variables. 

To learn the shared interaction pattern, we first randomly initialize a learnable $N$-dimensional memory embeddings $\mathcal{M}^{(i)} \in \mathbb {R}^{\bar{D} \times N}$ for each patch, with indices $i \in [0,N-1]$.  
As shown in Figure~\ref{fig:framework_LSTINet}(b), 
$\mathcal{M}^{(i)}$ is further encoded by an MLP encoder consisting of three linear layers with various hidden sizes, which enhances the learnability of the memory $\mathcal{M}^{(i)}$. 
During training, the model learns a Bernoulli distribution that can utilize $\mathcal{M}$ to construct the shared connection matrix $\mathbb{C}$ based on the training samples and gradient descent algorithm.
During inference, the input samples don't influence the generation of the shared connection matrix $\mathbb{C}$, which is generated independently from the well-learned memory $\mathcal{M}$ and Bernoulli distribution. We will proceed to explain how to learn the connection matrix $\mathbb{C}$.

\textbf{Rough Patch Connections Learning}
The connection probabilities from patch $i$ to patch $j$ are defined as $c_{ij}=\{ c^0_{ij},c^1_{ij}\}$ and $c^0_{ij}+c^1_{ij}=1$, which follows a standard Bernoulli distribution. 
$c^1_{ij}$ denotes the probability of existing potentially important connections between the pair of patches. 
$c_{ij}$ is an element in the connection matrix $\mathbb{C}$ with dimension $N\times N$. 
Each row of $\mathbb{C}$ represents $N$ pairs of patches. 
Taking the first row for example, the patch pairs are $\{c_{(0,0)},c_{(0,1)},\dots,c_{(0,N-1)}\}$. 
By using the row and column indices of all patch pairs in the first row, we obtain patch set $C_1^{\text{row}}= \{0,0,0,\dots,0\}$ ($N$ repeated patches) and patch set $C_1^{\text{col}}= \{0,1,2,\dots,N-1\}$. 
In $C_1^{\text{col}}$, subscript 1 indicates the first row, and ``col'' represents the indices of all columns in the current row.
All patch pairs in the first row are formed by combining patches from patch set $C_1^{\text{row}}$ and patch set $C_1^{\text{col}}$ in pairs.
Considering all rows of $\mathbb{C}$, we obtain $\mathbb{C}^{\text{row}}=\{C_1^{\text{row}},\dots,C_N^{\text{row}} \}$ and $\mathbb{C}^{\text{col}}=\{C_1^{\text{col}},\dots,C_N^{\text{col}} \}$. 
$\mathbb{C}^{\text{row}}$ and $\mathbb{C}^{\text{col}}$ respectively represent two sets of patches and they together form the patch pairs.
Our purpose is to find the effective connection between time patches set $\mathbb{C}^{\text{row}}$ and $\mathbb{C}^{\text{col}}$. 
Specifically, we use a neural network to predict the connection probability $c_{ij}$ that follows the Bernoulli distribution:
 \begin{equation}
\label{equ:features_e}
H=\mathcal{M}[\mathbb{C}^{\text{row}},:]\parallel\mathcal{M}[\mathbb{C}^{\text{col}},:]
% %\vspace{-0.1cm}
\end{equation}
% %\vspace{-0.2cm}
 \begin{equation}
\label{equ:features_e}
c=\Theta_2 ReLU(H\Theta_1+\textbf{b}_1)+\textbf{b}_2
% %\vspace{-0.1cm}
\end{equation}
% %\vspace{-0.2cm} 
where $\parallel$ denotes concatenate operation. $\Theta$ and \textbf{b} are learnable parameters.

Now, we can sample the discrete connection result $z_{ij} \in \{0,1\}$ based on the probability $c_{ij}$, which is obtained from the Bernoulli distribution. 
If $c_{ij}>0.5$, $ z_{ij}=1$ and this denotes existing potentially important connection between patch $i$ and $j$, vice versa. 
However, the discrete sampling is non-differentiable. 
The gradients obtained by prediction errors and backpropagation cannot flow through the learnable parameters involved in the connection prediction, leading the model can't learn the important connections between patches for capturing temporal dependencies. 
To address this, we first use the Gumbel-Max trick~\cite{DBLP:conf/iclr/MaddisonMT17,DBLP:conf/iclr/JangGP17} to make the sampling process of the discrete variable $z_{ij}$ differentiable:
\begin{equation}
% %\vspace{-0.2cm}
\label{equ:final_scale_factors}
    z_{ij}=\mathop{\arg\max}_{a \in \{0,1\}}\text{log} c^a_{ij}+g^a_{ij}
% %\vspace{-0.1cm}
\end{equation}
where $g_0$ and $g_1$ draw from a standard Gumbel distribution, which is sampled by drawing $u\sim$ Uniform(0,1) and computing $g=-\text{log}(-\text{log}u)$. 

Next, we further substitute the $\mathop{\arg\max}$ with the Gumbel-Softmax reparameterization trick~\cite{DBLP:conf/iclr/MaddisonMT17,DBLP:conf/iclr/JangGP17} for being differentiable:

% Due to the $\mathop{\arg\max}$ is still not differentiable, we further substitute it with the Gumbel-Softmax reparameterization trick~\cite{DBLP:conf/iclr/MaddisonMT17,DBLP:conf/iclr/JangGP17} as:
 \begin{equation}
 \label{equ:learn_connection}
  z_{ij} = \sum_{e=0}^1 \frac{exp((\text{log} c^a_{ij}+g^a_{ij})/\tau)}{\sum_{a\in \{0,1\}} exp((\text{log} c^a_{ij}+g^a_{ij})/\tau) }\times e
%\vspace{-0.1cm}
\end{equation}
where $\tau$ is the temperature parameter to control the Gumbel-Softmax distribution's smoothness. 
As the temperature $\tau$ approaches 0, the Gumbel-Softmax distribution becomes identical to the one-hot Bernoulli distribution. 

We can now learn the potentially important connections $z_{ij}$ between any pair of time patches by using standard gradient descent algorithms. 
However, due to the standard Bernoulli distribution not considering sparsity, the $N\times N$ connection matrix $\mathbb{C}$ learned by standard Bernoulli distribution can be redundant, which may introduce noise and harm the TSF accuracy. 
Hence, $\mathbb{C}$ learned by the standard Bernoulli distribution can be seen as merely a rough learning of the important connections between patches. 
We propose to use sparsity to induce the Bernoulli distribution to generate sparse connections between patches, thereby refinedly learning the more important connections. 
We will proceed to explain the sparsity design.

\textbf{self-Adaptive Sparse Regularization Loss (ASRL)}
We further propose a self-adaptive sparse regularization loss to add sparsity for the Bernoulli distribution, which forces the model to abandon lower effective connections and find the most important ones. 
Specifically, during the training process, we intermittently extract the top $K$ most important connections as one-hot labels for sparsity regularization and they are defined as $\mathbb{C}^{\text{top}}$. 
$K$ is computed by $K=N \times N \times \delta$, where $N$ is the total number of patches and 1-$\delta$ is the sparse rate (fixed at 0.15 in this paper).
The element values of top $K$ connections are set to 1 and others are 0.
$\mathbb{C}^{\text{top}}$ is used to calculate the cross-entropy loss with the predicted probabilities of all pair patches $\mathbb{C}^{\text{pre}}=\{c_{(0,0)},\dots,c_{(N-1,N-1)}\}$, providing self-adaptive sparse regularization for patch connections learning. 
ASRL can be formulated as follows:
\begin{equation}
%\vspace{-0.1cm}
\label{equ:sparse_loss}
    \mathcal{L}=\sum\nolimits_{ij}-\mathbb{C}^{\text{top}}_{ij}\text{log}\mathbb{C}^{\text{pre}}_{ij}-(1-\mathbb{C}^{\text{top}}_{ij})\text{log}(1-\mathbb{C}^{\text{pre}}_{ij})
% %\vspace{-0.1cm}
\end{equation}

\textbf{Proposed Two Training Strategies} 
For the first strategy, to prevent excessive sparsity constraints from leading to insufficient sparsity exploration, we apply sparsity regularization intermittently, i.e., only when the indicator $\widehat{e}_{gn}=1$:
% %\vspace{-0.1cm}
\begin{equation}
% %\vspace{-0.2cm}
\label{equ:cal_global_randomness}
    \widehat{e}_{gn}=
    \begin{cases}
    1 &  \text{if} \ \ \text{current epoch} \ \% \ \eta=0 \\
    0 &  \text{otherwise} \\
    \end{cases}
%\vspace{-0.1cm}
\end{equation}
where $\eta$ is used to control the frequency of adding ASRL.

For the second strategy, the number of $N$ determines the size of the connection matrix $\mathbb{C}$. 
The huge $\mathbb{C}$ will increase the model's convergence difficulty.
Therefore, we propose to limit the size of $\mathbb{C}$ as $\widetilde{N} \times \widetilde{N}$ by defining a rule for a different historical input $\mathcal{X}_h$ with length $n$. 
Specifically, in our method, $L=2K$, substituting this into the Eq.~\ref{equ:patch_cons} and simplifying, we obtain:
 \begin{equation}
   %\vspace{-0.1cm}
\label{equ:patch_rule}
\widetilde{K}=\lfloor n/\widetilde{N} \rfloor, \ \ \ \widetilde{L}= 2 \widetilde{K} 
 % %\vspace{-0.1cm}
\end{equation}

\textbf{Multi-head Interactions}
Through Eq.~\ref{equ:learn_connection}, we obtain $\mathbb{C}^z=\{z_{(0,0)}, \dots, z_{(N-1,N-1)}\}$, which contains most important connections between time patches with $z_{ij}=1$. 
Since the learned patch connections in $\mathbb{C}$ are static and independent of TS input during inference, this may lead to the failure of $\mathbb{C}$ because the patches in the input time window dynamically change. 
To address this, we propose a time-invariant component, which comprises an MLP layer to mix the patch feature of $\mathcal{X}_d$ into different patch positions and $\mathcal{X}_d$ undergoes the time-invariant component becoming $\bar{\mathcal{X}}_d$.
Now, for any input window, the model can capture temporal dependency through the learned shared connections between patches. 

Capturing temporal dependency is realized by Sparse Time Patch Propagation and Time patch Updating (Time Updating) in Figure~\ref{fig:framework_LSTINet}(a). 
Through time patch propagation, temporal information is propagated among patches, preparing for capturing temporal dependencies. 
This is finished based on multiplication between the shared connection matrix $\mathbb{C}$ and patch feature $V$ ($V=\bar{\mathcal{X}}_d \Theta_1+\textbf{b}_1$), as shown in the green rectangle of Figure~\ref{fig:framework_LSTINet}(a). 
We use multi-head time patch propagation to make the model generate diversely effective propagations, which can be formulated as:
\begin{equation}
% %\vspace{-0.2cm}
\label{equ:interaction_head}
    \widehat{\mathcal{X}}_d=\big( \mathbb{C}^z_1 \parallel \dots \parallel \mathbb{C}^z_h \big) \big(V_1 \parallel \dots \parallel V_h \big)
% %\vspace{-0.1cm}
\end{equation}
where $h$ is the total number of heads.

Then, after $V_h$ undergoes each propagation head, some patches in $V_h$ have received information propagated from other patches. 
We further update the temporal information of each patch through the Time Updating component, i.e., updating their representations for capturing temporal dependency, which is finished by using an MLP to project the temporal patch dimension of $\widehat{\mathcal{X}}_d$.

The Linear Align component in Figure~\ref{fig:framework_LSTINet} is used to align the feature dimension to enable the residual connections, which helps LSINet mitigate performance degradation when stacking STI modules. 
The MLP in the ``Integration'' part of Figure~\ref{fig:framework_LSTINet} is to integrate the interacted temporal features. 
Based on residual connections, the output of the STI module is $\widetilde{\mathcal{X}}_d$, which is flattened and sent to a linear predictor for forecasting. LSINet is formed by the stacked STI modules.

 \begin{table*}[bt]
 %\vspace{-0.3cm}
    % \renewcommand{\arraystretch}{0.5}
    \setlength{\tabcolsep}{2.7pt}
    % {|>{\setlength{\tabcolsep}{3pt}}c|c|c|}
    \centering
  
    %\vspace{-0.2cm}

    % \begin{tabular}{c|c|p{20pt}p{20pt}|cc|cc|cc|cc|cc}
    % \small
    { \small
    \begin{tabular}{c|c|cc|cc|cc|cc|cc|cc|cc|cc}
        \hline
        \multirow{2}{*}{\shortstack{}} & &  \multicolumn{2}{c|}{LSINet} &  \multicolumn{2}{c|}{CI-TSMixer}&  \multicolumn{2}{c|}{FiLM}&  \multicolumn{2}{c|}{DLinear} & \multicolumn{2}{c|}{PatchTST} & \multicolumn{2}{c|}{TimeMixer}& \multicolumn{2}{c|}{Scaleformer}& \multicolumn{2}{c}{Pathformer}\\
         & & MSE & MAE & MSE & MAE & MSE & MAE  & MSE & MAE & MSE & MAE& MSE & MAE& MSE & MAE& MSE & MAE \\ 
         \midrule[0.5pt]
         \multirow{4}{*}{\rotatebox[origin=c]{90}{ETTh1}} &96 &\textbf{0.366$\pm$2e-4} &\textbf{0.391$\pm$1e-6}&0.373 &0.398&\underline{0.371} &\underline{0.394}&0.375 &0.399&0.37 &0.4&0.375 &0.405&0.379 &0.409&0.393 &0.406\\
         &192 &\textbf{0.400$\pm$3e-4} &\textbf{0.412$\pm$2e-4}&0.405 &0.427&0.414 &0.423&0.405 &0.416&0.413 &0.429&\underline{0.408} &\underline{0.423}&0.411 &0.43&0.421 &0.436\\
         & 336 &\underline{0.427}$\pm$6e-4 &\textbf{0.428$\pm$3e-4}&0.426 &0.441&0.442 &0.445&0.439 &0.443&\textbf{0.422} &\underline{0.44}&0.435 &0.444&0.43 &0.443&0.451 &0.451\\
         & 720  &\textbf{0.441$\pm$2e-3} &\textbf{0.457$\pm$2e-3}&\underline{0.441} &\underline{0.46}&0.465 &0.472&0.472 &0.49&0.447 &0.468&0.457 &0.469&0.446 &0.465&0.483 &0.47\\
         \midrule[0.5pt]
        
        \multirow{4}{*}{\rotatebox[origin=c]{90}{ETTh2}} &96 &\textbf{0.267$\pm$7e-5} &\textbf{0.337$\pm$4e-4}&0.278 &0.344&0.284 &0.348&0.289 &0.353&\underline{0.274} &\underline{0.337}&0.286 &0.347&0.275 &0.343&0.285 &0.349\\
         &192 &\textbf{0.324$\pm$4e-4} &\textbf{0.372$\pm$8e-5}&0.338 &0.382&0.357 &0.4&0.383 &0.418&0.341 &0.382&0.347 &0.384&0.337 &0.384&\underline{0.331} &\underline{0.385}\\
         & 336 &\textbf{0.345$\pm$8e-4} &\textbf{0.394$\pm$9e-4}&\underline{0.356} &\underline{0.405}&0.377 &0.417&0.448 &0.465&0.359 &0.405&0.374 &0.41&0.364 &0.414&0.368 &0.409\\
         & 720 &\textbf{0.382$\pm$7e-4} &\underline{0.430}$\pm$6e-4&0.391 &0.433&0.439 &0.456&0.605 &0.551&\underline{0.388} &\textbf{0.427}&0.406 &0.44&0.397 &0.438&0.389 &0.427\\
         \midrule[0.5pt]
        
        \multirow{4}{*}{\rotatebox[origin=c]{90}{ETTm1}} &96 &\textbf{0.293$\pm$0.001} &\textbf{0.341$\pm$8e-4}&\underline{0.293} &\underline{0.343}&0.302 &0.349&0.305 &0.353&0.297 &0.348&0.295 &0.35&0.293 &0.347&0.301 &0.352\\
         &192 &\textbf{0.329$\pm$4e-4} &\textbf{0.363$\pm$5e-4}&\underline{0.331} &\underline{0.365}&0.338 &0.373&0.33 &0.369&0.333 &0.376&0.332 &0.369&0.333 &0.371&0.356 &0.383\\
         & 336 &\textbf{0.357$\pm$8e-4} &\textbf{0.384$\pm$7e-4}&0.363 &0.384&0.365 &0.385&\underline{0.36} &\underline{0.384}&0.359 &0.392&0.365 &0.391&0.364 &0.391&0.387 &0.405\\
         & 720 &\textbf{0.396$\pm$9e-4} &\textbf{0.409$\pm$8e-4}&0.405 &0.419&0.42 &0.42&0.405 &0.413&\underline{0.397} &\underline{0.42}&0.416 &0.424&0.42 &0.425&0.416 &0.42\\
         \midrule[0.5pt]

        \multirow{4}{*}{\rotatebox[origin=c]{90}{ETTm2}} &96&\textbf{0.161$\pm$5e-4} &\textbf{0.256$\pm$6e-4}&0.166 &0.258&0.165 &0.256&0.163 &0.259&\underline{0.163} &\underline{0.258}&0.169 &0.261&0.172 &0.255&0.168 &0.258\\
         &192 &\textbf{0.213$\pm$0.001} &\textbf{0.295$\pm$0.001}&0.222 &0.296&0.222 &0.296&0.218 &0.302&\underline{0.216} &\underline{0.296}&0.227 &0.3&0.231 &0.298&0.227 &0.299\\
         & 336 &\textbf{0.257$\pm$0.001} &\textbf{0.327$\pm$0.001}&0.274 &\underline{0.328}&0.277 &0.333&0.27 &0.34&\underline{0.266} &0.329&0.274 &0.329&0.276 &0.328&0.273 &0.331\\
         & 720 &\textbf{0.328$\pm$0.002} &\textbf{0.372$\pm$3e-4}&0.34 &0.38&0.371 &0.389&0.368 &0.406&\underline{0.339} &\underline{0.379}&0.352 &0.384&0.349 &0.383&0.366 &0.392 \\
         \midrule[0.5pt]
      
        \multirow{4}{*}{\rotatebox[origin=c]{90}{Weather}} &96 &\textbf{0.147$\pm$4e-4} &0.199$\pm$3e-4&\underline{0.147} &\textbf{0.197}&0.199 &0.262&0.165 &0.224&0.149 &0.198&0.146 &\underline{0.198}&0.152 &0.208&0.155 &0.208\\
         &192 &\underline{0.192}$\pm$8e-4 &\underline{0.243}$\pm$0.001&0.192 &0.24&0.228 &0.288&0.207 &0.263&0.194 &0.241&\textbf{0.190} &\textbf{0.240}&0.197 &0.251&0.196 &0.246\\
         & 336 &\textbf{0.238$\pm$0.003} &0.281$\pm$0.003&\underline{0.243} &\underline{0.279}&0.267 &0.323&0.249 &0.294&0.244 &0.282&0.242 &0.283&0.253 &0.296&0.25 &0.286\\
         & 720 &\textbf{0.304$\pm$0.002} &\textbf{0.332$\pm$0.002}&0.31 &0.333&0.319 &0.361&0.308 &0.344&\underline{0.307} &\underline{0.33}&0.314 &0.333&0.311 &0.343&0.324 &0.337\\
         \midrule[0.5pt]

        \multirow{4}{*}{\rotatebox[origin=c]{90}{Electricity}} &96 &\textbf{0.129$\pm$1e-4} &\textbf{0.224$\pm$6e-5}&\underline{0.131} &\underline{0.226}&0.154 &0.267&0.133 &0.232&0.135 &0.231&0.133 &0.229&0.143 &0.247&0.145 &0.236\\
         &192 &\textbf{0.144$\pm$2e-4} &\textbf{0.240$\pm$2e-4}&\underline{0.147} &\underline{0.244}&0.164 &0.258&0.15 &0.249&0.15 &0.244&0.15 &0.245&0.161 &0.266&0.167 &0.256\\
         & 336 &\textbf{0.158$\pm$3e-4} &\textbf{0.256$\pm$3e-4}&\underline{0.163} &\underline{0.261}&0.188 &0.283&0.165 &0.267&0.166 &0.261&0.168 &0.264&0.179 &0.285&0.186 &0.275\\
         & 720 &\textbf{0.192$\pm$8e-4} &\textbf{0.287$\pm$4e-4}&\underline{0.200} &\underline{0.294}&0.236 &0.332&0.2 &0.301&0.206 &0.294&0.205 &0.296&0.214 &0.318&0.231 &0.309\\

        \midrule[0.5pt]
    \end{tabular}}%small
      \caption{Long-term TSF task on public datasets over different baselines. The advanced transformer models include PatchTST, Scaleformer, and Pathformer. The remaining are advanced linear models. 
     }
         \label{tab:metric_public_long_term}
%\vspace{-0.5cm}
\end{table*}

%\vspace{-0.1cm}
\section{Experiments and Results}
\subsection{Experimental Settings}
\textbf{Datasets and Evaluation Metrics}
We evaluate the performance of the proposed LSINet on 6 popular datasets, including Weather, Electricity, and 4 ETT datasets, covering a range of time steps (17420 to 69680) and variables (7 to 321) and have been widely employed in the literature for multivariate forecasting tasks~\cite{nie2022time_patchformer,wu2021autoformer,zhou2022fedformer}. 
Following previous works, we evaluate multivariate TSF tasks using Mean Squared Error (MSE) and Mean Absolute Error (MAE) metrics.

\textbf{Baselines}
We compare LSINet against the advanced linear and transformer models.
\textbf{(i) is advanced linear models, }including TimeMixer~\cite{wangtimemixer}, CI-TSMixer~\cite{vijay2023tsmixer}, FiLM~\cite{zhou2022film}, and DLinear~\cite{zeng2023transformers_linear}.
\textbf{(ii) is advanced transformer models, }including PatchTST~\cite{nie2022time_patchformer}, Pathformer~\cite{chen2024pathformer} and Scaleformer~\cite{shabani2022scaleformer}. Since Scaleformer is a general architecture, we combine it with Autoformer~\cite{wu2021autoformer}, NHits~\cite{challu2022n_Nhits}, and PatchTST to create strong baselines.

%  \begin{table} [bt] %[h!]
%     % \renewcommand{\arraystretch}{0.5}
%     \setlength{\tabcolsep}{2.5pt}
%     % {|>{\setlength{\tabcolsep}{3pt}}c|c|c|}
%     \centering
%     %\vspace{-0.2cm}
%     \caption{Statistics of used popular public datasets.}
%     %\vspace{-0.3cm}
%     \label{tab:dataset_stat}
%     % \begin{tabular}{c|c|p{20pt}p{20pt}|cc|cc|cc|cc|cc}
%     { \small
%     \begin{tabular}{c|c|c|c|c|c|c}
%         \hline
%         \multirow{1}{*}{\shortstack{Datasets}}  &  \multicolumn{1}{c|}{ETTh1} &  \multicolumn{1}{c|}{ETTm1}  & \multicolumn{1}{c|}{Weather}& \multicolumn{1}{c|}{ETTh2}& \multicolumn{1}{c|}{ETTm2}& \multicolumn{1}{c}{Electricity}\\
%          \midrule[0.5pt]
%          \multirow{1}{*}{Features}  &7 &7 &21 &7&7&321\\
%         \midrule[0.5pt]
%          \multirow{1}{*}{Timesteps}  &17420 &69680 &52696&17420&69680&26304\\
%         \midrule[0.5pt]
%     \end{tabular}
%     } % \small
% %\vspace{-0.5cm}
% \end{table}

\subsubsection{Implementation Details}
For LSINet, the hidden size for patch embedding, position embedding (Eq.~\ref{equ:linear_proj_pos}), and all used MLPs are fixed at 128. The multi-head $h$ is fixed at 4.
$\eta \in \{1,3\}$ is used for controlling the interval of using sparse regularization loss. 
The number of patch $\widetilde{N}$ is fixed at 64 for sparse interaction learning. 
$\delta$ for controlling sparsity is fixed at 0.15, i.e., the sparse rate of $\mathbb{C}$ is 0.85.
The number of stacked STI modules is fixed at 1 on all datasets. 
In TSF tasks on different datasets, we don't search or adjust the above hyper-parameters.
Following the previous works~\cite{zhou2021informer,zhou2022film,challu2022n_Nhits,das2023long}, the input length $n \in \{256,384, 512, 768,1024, 1536, 2048\}$ is tuned for best forecasting performance, which is also tuned for other comparison methods to create strong baselines.  
The prediction lengths are fixed to be 96, 192, 336, and 720, respectively. 
The learning rate is fixed at 1e-4. 
The batch size for 4 ETT datasets is fixed at 128 while for Weather and Electricity datasets are fixed at 64 and 32 respectively. 
All methods follow the same data loading parameters (e.g., train/val/test split ratio) as in~\cite{nie2022time_patchformer}. 
We will show more details in our full version of the paper due to limited space.
For each experiment, we independently ran 5 times with 5 different seeds for 30 epochs and reported the average metrics and standard deviations. Experiments are conducted on NVIDIA GeForce RTX 3090 GPU on PyTorch.

%\vspace{-0.1cm}
\subsection{Time Series Forecasting (TSF) Accuracy}
The whole results are shown in Table~\ref{tab:metric_public_long_term}.
Since we observe similar conclusions in MSE and MAE, we explain all the results in this paper using the MSE metric. 
For comparisons, we average the
maximum and average improvement ratios of our method over baselines across all prediction lengths and datasets.
For overall \textit{average}\&\textit{maximum} improvement ratios over MSE across all prediction steps and datasets, LSINet not only achieves improvements over advanced linear models (CI-TSMixer: 2.06\%\&3.40\%, FiLM: 8.62\%\&13.37\%, DLiner: 6.61\%\&12.19\% and TimeMixer: 3.7\%\&5.4\%) but also over advanced transformer models (PatchTST: 2.22\%\&3.55\%, Scaleformer: 4.73\%\&6.32\%, and Pathformer: 6.84\%\&9.37\%). 
Moreover, LSINet achieves these improvements with a significant efficiency improvement and we will discuss it in Table~\ref{tab:Memory_usage}.
%\vspace{-0.1cm}
\subsection{Time Series Forecasting Efficiency}
\label{sec:comp_efficiency}
\begin{figure}[bt]
\centerline{\includegraphics[width=1.0\linewidth]{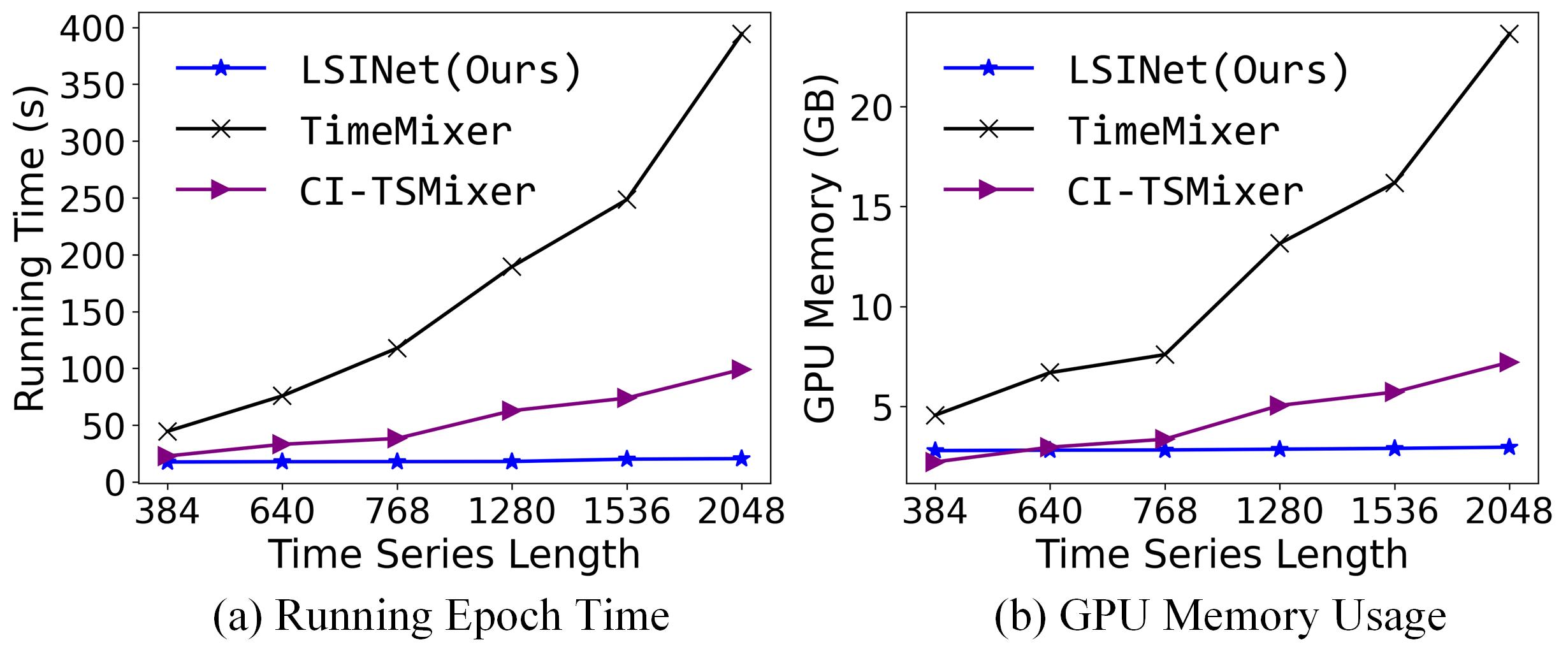} }
%\vspace{-0.4cm}
\caption{
Efficiency analysis for best-performing linear models on Weather dataset (batch size 64) with training epoch time (a) and GPU memory usage (b).}
%\vspace{-0.2cm}
\label{fig:speed_memory_0531}
\end{figure}

Since CI-TSMixer has been demonstrated significantly faster and uses less memory than PatchTST~\cite{vijay2023tsmixer}, we don't include it in the efficiency comparison. 
Here, we don't adjust the hyperparameters of each model used in Table~\ref{tab:metric_public_long_term}, but only change the input TS length to compare the efficiency of different methods.
Comparison results in Figure~\ref{fig:speed_memory_0531} show that LSINet achieves significantly higher running efficiency and lower memory usage than 
advanced linear models CI-TSMixer and TimeMixer. 

\begin{table}[bt]
% %\vspace{-0.15cm}
    % \renewcommand{\arraystretch}{0.5}
    \setlength{\tabcolsep}{1.5pt}
    % {|>{\setlength{\tabcolsep}{3pt}}c|c|c|}
    \centering
    
    %\vspace{-0.2cm}

    % \begin{tabular}{c|c|p{20pt}p{20pt}|cc|cc|cc|cc|cc}
    % \small
    { \small
    \begin{tabular}{c|c|c|c|c|c|c}
        \hline
        \multirow{1}{*}{\shortstack{}} & &  \multicolumn{1}{c|}{ETTh1} &  \multicolumn{1}{c|}{ETTh2}&  \multicolumn{1}{c|}{ETTm1}  & \multicolumn{1}{c|}{ETTm2} & \multicolumn{1}{c}{Weather}\\
         % & & MSE  & MSE  & MSE  & MSE \\ 
         \midrule[0.5pt]
         \multirow{4}{*}{\shortstack{LSINet}} &96 &\textbf{0.366}&\textbf{0.267}&0.293&\textbf{0.161}&\textbf{0.147}\\
         &192 &\textbf{0.400}&\textbf{0.324}&\textbf{0.329}&\textbf{0.213}&0.192\\
         & 336 &\textbf{0.427}&\textbf{0.345}&\textbf{0.357}&\textbf{0.257}&0.238 \\
         & 720  &\textbf{0.441}&\textbf{0.382}&\textbf{0.396}&\textbf{0.328}&0.304 \\
        \midrule[0.5pt]
        \multirow{4}{*}{\shortstack{w/o MSIM\\(MSE increases\\2.23\% \& 6.54\% \\on 
        Avg. \& Max.)}} &96 &0.367&0.273&0.296&0.163&0.164\\
         &192 &0.402&0.331&0.334&0.215&0.205 \\
         & 336 &0.427&0.358&0.358&0.259&0.251 \\
         & 720  &0.443&0.391&0.398&0.334&0.307 \\
        \midrule[0.5pt]
        \multirow{4}{*}{\shortstack{Using SAM\\(MSE increases\\2.24\% \& 6.53\% \\on 
        Avg. \& Max.)}} &96 &0.369&0.274&\textbf{0.291}&0.167&0.148\\
         &192 &0.407&0.327&0.330&0.224&\textbf{0.190} \\
         & 336 &0.436&0.349&0.364&0.276&\textbf{0.236} \\
         & 720  &0.451&0.383&0.435&0.353&\textbf{0.300}\\
        \midrule[0.5pt]
    \end{tabular}}
    \caption{Ablation study of not using Multihead Sparse Interaction Mechanism (w/o MSIM) and replacing MSIM with Self-Attention Mechanism (SAM).
    MSE increases of Avg. and Max. are computed by averaging the average and max increases on all prediction lengths and five public datasets.}
        \label{tab:ablation_study}
%\vspace{-0.5cm}
\end{table}

\begin{table}[bt]
    \setlength{\tabcolsep}{1.8pt}
    % {|>{\setlength{\tabcolsep}{3pt}}c|c|c|}
    \centering
    % %\vspace{-0.2cm}
   
    %\vspace{-0.2cm}
    % \begin{tabular}{c|c|p{20pt}p{20pt}|cc|cc|cc|cc|cc}
    % \small

    { \small
    \begin{tabular}{cc|c|c|c|c|c|c|c|c} 
        \hline
      & &  \multicolumn{4}{c|}{LSINet} &  \multicolumn{4}{c}{LSINet w/o ASRL}\\
         & &96&192&336&720 &96&192&336&720   \\ 
         \midrule[0.5pt]

         &Weather &0.147 &0.192 &0.238 &0.304 &0.150 &0.198 &0.246 &0.307\\
        \midrule[0.5pt]
         &Electricity &0.129 &0.144 &0.158 &0.192 &0.129&0.145&0.160&0.194\\
         
        % \midrule[0.5pt]
        % \multicolumn{6}{c|}{\shortstack{LSINet \% average improvements}} &\multicolumn{1}{c|}{\shortstack{1.0}}&\multicolumn{1}{c|}{\shortstack{1.86}} &\multicolumn{1}{c|}{\shortstack{2.25}}&\multicolumn{1}{c}{\shortstack{1.0}}

        % \\
        \midrule[0.5pt]
    \end{tabular}
    } % \small
     \caption{Impact of not using self-Adaptive Sparse Regularization Loss (w/o ASRL) in Eq.~\ref{equ:sparse_loss}.}
         \label{tab:wo_asrl}
%\vspace{-0.5cm}
\end{table}

\subsection{Effectiveness of Multi-head Sparse Interaction Mechanism (MSIM)}
\textbf{Results of removing MSIM}.
As shown in Table~\ref{tab:ablation_study}, after removing the designed MSIM, we observe the MSE increases by 2.23\% and 6.54\% for average and maximum respectively. 
This indicates that MSIM is effective for capturing complex temporal dependence and improving TSF accuracy.

\textbf{Results of replacing MSIM with Self-Attention Mechanism (SAM)}.
\textbf{For accuracy,} 
we observe the MSE overall increases by 2.24\% and 6.53\% for average and maximum respectively, as shown in Table~\ref{tab:ablation_study}.
\textbf{For efficiency,} the training epoch time, inference time, and memory usage of using SAM is significantly higher than using MSIM, as shown in Tabel~\ref{tab:Memory_usage}. 
Comparison results suggest that SAM may not be well-suited for lightweight linear model architectures, but our MSIM shows better adaptability. 

\textbf{Results of not using self-Adaptive Sparse Regularization Loss (ASRL)}. As shown in Table~\ref{tab:wo_asrl}, after removing ASRL, the MSE averagely increases by 1\%, 1.86\%, 2.25\% and 1\% on forecasting 96,192,336 and 720 future values of two large datasets. 
This indicates that ASRL can force the
model to find the
most important connections between time patches and abandon lower effective ones by controlling the sparsity strength of the connection matrix $\mathbb{C}$.
Consequently, ASRL mitigates accuracy deterioration caused by introducing redundancy and noise.

 \begin{table}[bt]
 %\vspace{-0.1cm}
    % \renewcommand{\arraystretch}{0.5}
    \setlength{\tabcolsep}{4pt}
    % {|>{\setlength{\tabcolsep}{3pt}}c|c|c|}
    \centering

    %\vspace{-0.2cm}
    % \begin{tabular}{c|c|p{20pt}p{20pt}|cc|cc|cc|cc|cc}
    % \small
    \small{
    \begin{tabular}{c|ccc|ccc} 
        \hline
        \multirow{2}{*}{\shortstack{Datasets/\\ Methods}} &  \multicolumn{3}{c|}{LSINet} &  \multicolumn{3}{c}{LSINet-SAM}\\
         &  \textit{Epo.$^T$} &\textit{Infer$^T$}& \textit{Mem.}  &  \textit{Epo.$^T$} &\textit{Infer$^T$}  & 
         \textit{Mem.}    \\ 
        \midrule[0.5pt]

        \multicolumn{1}{c|}{Weather} &22.36 &20.50 &3026 &56.45  &50.59 &7584\\
        \midrule[0.5pt]
        \multicolumn{1}{c|}{Electricity} &91.25 &82.53 &6936 &291.47  &175.13 &23464\\
        \midrule[0.5pt]

    \end{tabular}}
    \caption{Efficiency comparison of LSINet using MSIM  and SAM on training time \textit{Epo.$^T$}(s/epoch), inference time \textit{Infer$^T$}(s) and memory \textit{Mem.}(MB) with input length 1024.}
        \label{tab:Memory_usage}
%\vspace{-0.7cm}
\end{table}

\begin{figure}[bt]
\centerline{\includegraphics[width=\linewidth]{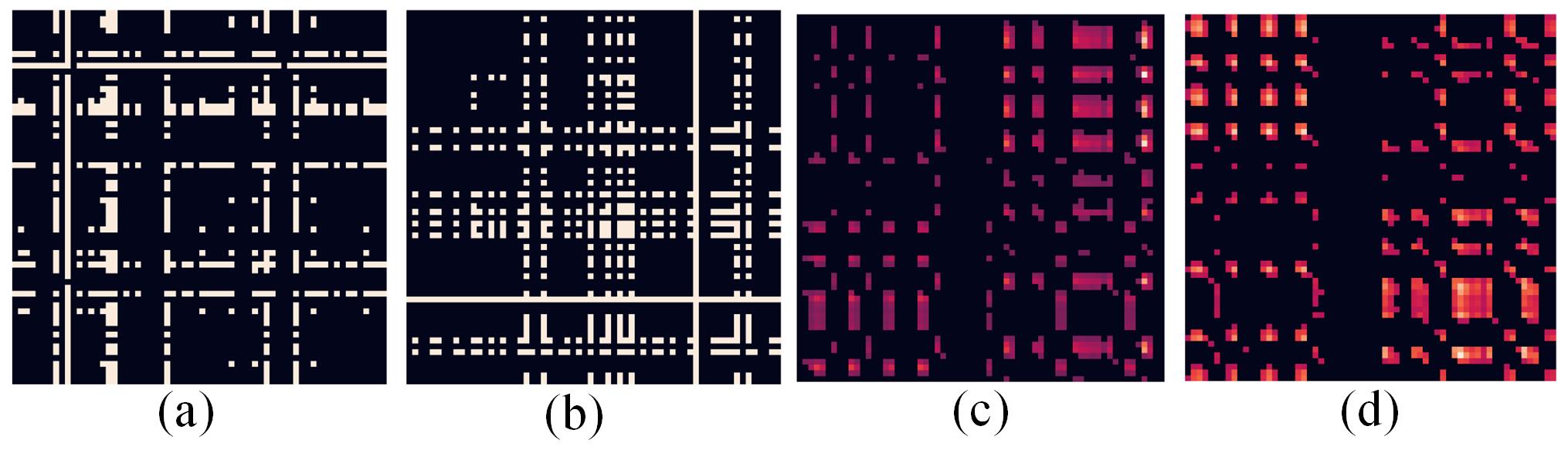} }
%\vspace{-0.4cm}
\caption{Learned shared sparse connection matrices by LSINet: (a)-(b) and learned self-attention matrices by PatchTST: (c)-(d) on the ETTm1 dataset. Both of them exhibit patterns of scattered points and blocks. }
%\vspace{-0.5cm}
\label{fig:SSCL_comp_SAM}
\end{figure}
%\vspace{0.1cm}
\subsection{Visualization of Learned Shared Sparse Interaction}
%\vspace{-0.1cm}
We observe the learned heatmaps of shared sparse interaction matrix $\mathbb{C}$  with sparse rate 0.85 (Figure~\ref{fig:SSCL_comp_SAM}(a)-(b)) by LSINet are similar to the learned heatmaps from self-attention mechanism (85\% lower scores are set to zero) by PatchTST model.
Both of them exhibit patterns of scattered points and blocks in heatmaps. 
This indicates that the Shared Sparse Connection Learning (SSCL) module can indeed learn effective temporal interactions that are captured by the self-attention mechanism and help improve the accuracy of TSF based on the lightweight linear model.

\section{Conclusion}
We propose a Lightweight Sparse Interaction Network
(LSINet) for TSF. 
Key designs of LSINet are Multihead Sparse Interaction
Mechanism (MSIM) and Shared Interactions Learning (SIL).
Instead of computing self-attention scores, LSINet directly learns important connections between time steps for capturing temporal dependence through the sparsity-induced
Bernoulli distribution (realized by sparsity regularization loss). 
Extensive experiments demonstrate that MSIM achieves better accuracy and efficiency than Self-Attention Mechanism (SAM) when combined with the linear model structure.
LSINet also achieves superior accuracy and efficiency over advanced TSF baselines.

\section{Acknowledgements}
This work is supported by the Ministry of Science and Technology of China, National Key Research and Development Program (No. 2021YFB3300503). 

\bibliography{aaai25}

\begin{thebibliography}{28}
\providecommand{\natexlab}[1]{#1}

\bibitem[{Alnegheimish et~al.(2024)Alnegheimish, Nguyen, Berti-Equille, and Veeramachaneni}]{alnegheimish2024large}
Alnegheimish, S.; Nguyen, L.; Berti-Equille, L.; and Veeramachaneni, K. 2024.
\newblock Large language models can be zero-shot anomaly detectors for time series?
\newblock \emph{arXiv preprint arXiv:2405.14755}.

\bibitem[{Angryk et~al.(2020)Angryk, Martens, Aydin, Kempton, Mahajan, Basodi, Ahmadzadeh, Cai, Filali~Boubrahimi, Hamdi et~al.}]{angryk2020multivariate_intr_weather}
Angryk, R.~A.; Martens, P.~C.; Aydin, B.; Kempton, D.; Mahajan, S.~S.; Basodi, S.; Ahmadzadeh, A.; Cai, X.; Filali~Boubrahimi, S.; Hamdi, S.~M.; et~al. 2020.
\newblock Multivariate time series dataset for space weather data analytics.
\newblock \emph{Scientific data}, 7(1): 227.

\bibitem[{Challu et~al.(2022)Challu, Olivares, Oreshkin, Garza, Mergenthaler, and Dubrawski}]{challu2022n_Nhits}
Challu, C.; Olivares, K.; Oreshkin, B.; Garza, F.; Mergenthaler, M.; and Dubrawski, A. 2022.
\newblock N-hits: Neural hierarchical interpolation for time series forecasting. arXiv.
\newblock \emph{arXiv preprint arXiv:2201.12886}.

\bibitem[{Chen et~al.(2001)Chen, Petty, Skabardonis, Varaiya, and Jia}]{chen2001freeway_traffic}
Chen, C.; Petty, K.; Skabardonis, A.; Varaiya, P.; and Jia, Z. 2001.
\newblock Freeway performance measurement system: mining loop detector data.
\newblock \emph{Transportation Research Record}, 1748(1): 96--102.

\bibitem[{Chen et~al.(2024)Chen, Zhang, Cheng, Shu, Wang, Wen, Yang, and Guo}]{chen2024pathformer}
Chen, P.; Zhang, Y.; Cheng, Y.; Shu, Y.; Wang, Y.; Wen, Q.; Yang, B.; and Guo, C. 2024.
\newblock Pathformer: Multi-scale transformers with Adaptive Pathways for Time Series Forecasting.
\newblock \emph{arXiv preprint arXiv:2402.05956}.

\bibitem[{Chen et~al.(2023)Chen, Li, Yoder, Arik, and Pfister}]{chen2023tsmixer}
Chen, S.-A.; Li, C.-L.; Yoder, N.; Arik, S.~O.; and Pfister, T. 2023.
\newblock Tsmixer: An all-mlp architecture for time series forecasting.
\newblock \emph{arXiv preprint arXiv:2303.06053}.

\bibitem[{Churpek, Adhikari, and Edelson(2016)}]{churpek2016value_intr_medical}
Churpek, M.~M.; Adhikari, R.; and Edelson, D.~P. 2016.
\newblock The value of vital sign trends for detecting clinical deterioration on the wards.
\newblock \emph{Resuscitation}, 102: 1--5.

\bibitem[{Das et~al.(2023)Das, Kong, Leach, Mathur, Sen, and Yu}]{das2023long}
Das, A.; Kong, W.; Leach, A.; Mathur, S.~K.; Sen, R.; and Yu, R. 2023.
\newblock Long-term Forecasting with TiDE: Time-series Dense Encoder.
\newblock \emph{Transactions on Machine Learning Research}.

\bibitem[{Das et~al.(2024)Das, Kong, Sen, and Zhou}]{dasdecoder}
Das, A.; Kong, W.; Sen, R.; and Zhou, Y. 2024.
\newblock A decoder-only foundation model for time-series forecasting.
\newblock In \emph{Forty-first International Conference on Machine Learning}.

\bibitem[{Ekambaram et~al.(2023)Ekambaram, Jati, Nguyen, Sinthong, and Kalagnanam}]{vijay2023tsmixer}
Ekambaram, V.; Jati, A.; Nguyen, N.; Sinthong, P.; and Kalagnanam, J. 2023.
\newblock TSMixer: Lightweight MLP-Mixer Model for Multivariate Time Series Forecasting.
\newblock In Singh, A.~K.; Sun, Y.; Akoglu, L.; Gunopulos, D.; Yan, X.; Kumar, R.; Ozcan, F.; and Ye, J., eds., \emph{Proceedings of the 29th {ACM} {SIGKDD} Conference on Knowledge Discovery and Data Mining, {KDD} 2023, Long Beach, CA, USA, August 6-10, 2023}, 459--469. {ACM}.

\bibitem[{Jang, Gu, and Poole(2017)}]{DBLP:conf/iclr/JangGP17}
Jang, E.; Gu, S.; and Poole, B. 2017.
\newblock Categorical Reparameterization with Gumbel-Softmax.
\newblock In \emph{5th International Conference on Learning Representations, {ICLR} 2017, Toulon, France, April 24-26, 2017, Conference Track Proceedings}. OpenReview.net.

\bibitem[{Jin et~al.(2023)Jin, Wang, Ma, Chu, Zhang, Shi, Chen, Liang, Li, Pan et~al.}]{jin2023time}
Jin, M.; Wang, S.; Ma, L.; Chu, Z.; Zhang, J.~Y.; Shi, X.; Chen, P.-Y.; Liang, Y.; Li, Y.-F.; Pan, S.; et~al. 2023.
\newblock Time-LLM: Time Series Forecasting by Reprogramming Large Language Models.
\newblock In \emph{The Twelfth International Conference on Learning Representations}.

\bibitem[{Kim et~al.(2022)Kim, Kim, Tae, Park, Choi, and Choo}]{DBLP:conf/iclr/KimKTPCC22}
Kim, T.; Kim, J.; Tae, Y.; Park, C.; Choi, J.; and Choo, J. 2022.
\newblock Reversible Instance Normalization for Accurate Time-Series Forecasting against Distribution Shift.
\newblock In \emph{The Tenth International Conference on Learning Representations, {ICLR} 2022, Virtual Event, April 25-29, 2022}. OpenReview.net.

\bibitem[{Li et~al.(2019)Li, Jin, Xuan, Zhou, Chen, Wang, and Yan}]{li2019enhancing}
Li, S.; Jin, X.; Xuan, Y.; Zhou, X.; Chen, W.; Wang, Y.-X.; and Yan, X. 2019.
\newblock Enhancing the locality and breaking the memory bottleneck of transformer on time series forecasting.
\newblock \emph{Advances in neural information processing systems}, 32.

\bibitem[{Liu et~al.(2024)Liu, Guo, Dai, Li, Bao, Ren, Jiang, and Xia}]{liu2024taming}
Liu, P.; Guo, H.; Dai, T.; Li, N.; Bao, J.; Ren, X.; Jiang, Y.; and Xia, S.-T. 2024.
\newblock Taming Pre-trained LLMs for Generalised Time Series Forecasting via Cross-modal Knowledge Distillation.
\newblock \emph{arXiv preprint arXiv:2403.07300}.

\bibitem[{Maddison, Mnih, and Teh(2017)}]{DBLP:conf/iclr/MaddisonMT17}
Maddison, C.~J.; Mnih, A.; and Teh, Y.~W. 2017.
\newblock The Concrete Distribution: {A} Continuous Relaxation of Discrete Random Variables.
\newblock In \emph{5th International Conference on Learning Representations, {ICLR} 2017, Toulon, France, April 24-26, 2017, Conference Track Proceedings}. OpenReview.net.

\bibitem[{Nie et~al.(2023)Nie, Nguyen, Sinthong, and Kalagnanam}]{nie2022time_patchformer}
Nie, Y.; Nguyen, N.~H.; Sinthong, P.; and Kalagnanam, J. 2023.
\newblock A Time Series is Worth 64 Words: Long-term Forecasting with Transformers.
\newblock In \emph{The Eleventh International Conference on Learning Representations, {ICLR} 2023, Kigali, Rwanda, May 1-5, 2023}. OpenReview.net.

\bibitem[{Shabani et~al.(2023)Shabani, Abdi, Meng, and Sylvain}]{shabani2022scaleformer}
Shabani, M.~A.; Abdi, A.~H.; Meng, L.; and Sylvain, T. 2023.
\newblock Scaleformer: Iterative Multi-scale Refining Transformers for Time Series Forecasting.
\newblock In \emph{The Eleventh International Conference on Learning Representations, {ICLR} 2023, Kigali, Rwanda, May 1-5, 2023}. OpenReview.net.

\bibitem[{Tan et~al.(2024)Tan, Merrill, Gupta, Althoff, and Hartvigsen}]{tan2024language}
Tan, M.; Merrill, M.~A.; Gupta, V.; Althoff, T.; and Hartvigsen, T. 2024.
\newblock Are Language Models Actually Useful for Time Series Forecasting?
\newblock \emph{arXiv preprint arXiv:2406.16964}.

\bibitem[{Vaswani et~al.(2017)Vaswani, Shazeer, Parmar, Uszkoreit, Jones, Gomez, Kaiser, and Polosukhin}]{vaswani2017attention}
Vaswani, A.; Shazeer, N.; Parmar, N.; Uszkoreit, J.; Jones, L.; Gomez, A.~N.; Kaiser, {\L}.; and Polosukhin, I. 2017.
\newblock Attention is all you need.
\newblock \emph{Advances in neural information processing systems}, 30.

\bibitem[{Wang et~al.(2024)Wang, Wu, Shi, Hu, Luo, Ma, Zhang, and Zhou}]{wangtimemixer}
Wang, S.; Wu, H.; Shi, X.; Hu, T.; Luo, H.; Ma, L.; Zhang, J.~Y.; and Zhou, J. 2024.
\newblock TimeMixer: Decomposable Multiscale Mixing for Time Series Forecasting.
\newblock In \emph{The Twelfth International Conference on Learning Representations}.

\bibitem[{Wu et~al.(2021)Wu, Xu, Wang, and Long}]{wu2021autoformer}
Wu, H.; Xu, J.; Wang, J.; and Long, M. 2021.
\newblock Autoformer: Decomposition transformers with auto-correlation for long-term series forecasting.
\newblock \emph{Advances in Neural Information Processing Systems}, 34: 22419--22430.

\bibitem[{Zeng et~al.(2023)Zeng, Chen, Zhang, and Xu}]{zeng2023transformers_linear}
Zeng, A.; Chen, M.; Zhang, L.; and Xu, Q. 2023.
\newblock Are transformers effective for time series forecasting?
\newblock In \emph{Proceedings of the AAAI conference on artificial intelligence}, volume~37, 11121--11128.

\bibitem[{Zhang et~al.(2024)Zhang, Huang, Wu, Lu, Qi, Chen, Xue, Wang, and Wang}]{zhang2024self}
Zhang, X.; Huang, Z.; Wu, Y.; Lu, X.; Qi, E.; Chen, Y.; Xue, Z.; Wang, P.; and Wang, W. 2024.
\newblock Self-Adaptive Scale Handling for Forecasting Time Series with Scale Heterogeneity.
\newblock In \emph{ICASSP 2024-2024 IEEE International Conference on Acoustics, Speech and Signal Processing (ICASSP)}, 7485--7489. IEEE.

\bibitem[{Zhou et~al.(2021)Zhou, Zhang, Peng, Zhang, Li, Xiong, and Zhang}]{zhou2021informer}
Zhou, H.; Zhang, S.; Peng, J.; Zhang, S.; Li, J.; Xiong, H.; and Zhang, W. 2021.
\newblock Informer: Beyond efficient transformer for long sequence time-series forecasting.
\newblock In \emph{Proceedings of the AAAI conference on artificial intelligence}, volume~35, 11106--11115.

\bibitem[{Zhou et~al.(2022{\natexlab{a}})Zhou, Ma, Wen, Sun, Yao, Yin, Jin et~al.}]{zhou2022film}
Zhou, T.; Ma, Z.; Wen, Q.; Sun, L.; Yao, T.; Yin, W.; Jin, R.; et~al. 2022{\natexlab{a}}.
\newblock Film: Frequency improved legendre memory model for long-term time series forecasting.
\newblock \emph{Advances in Neural Information Processing Systems}, 35: 12677--12690.

\bibitem[{Zhou et~al.(2022{\natexlab{b}})Zhou, Ma, Wen, Wang, Sun, and Jin}]{zhou2022fedformer}
Zhou, T.; Ma, Z.; Wen, Q.; Wang, X.; Sun, L.; and Jin, R. 2022{\natexlab{b}}.
\newblock Fedformer: Frequency enhanced decomposed transformer for long-term series forecasting.
\newblock In \emph{International Conference on Machine Learning}, 27268--27286. PMLR.

\bibitem[{Zhou et~al.(2023)Zhou, Niu, Sun, Jin et~al.}]{zhou2023one}
Zhou, T.; Niu, P.; Sun, L.; Jin, R.; et~al. 2023.
\newblock One fits all: Power general time series analysis by pretrained lm.
\newblock \emph{Advances in neural information processing systems}, 36: 43322--43355.

\end{thebibliography}

\end{document}